*Gene expression*

# A feasible roadmap for unsupervised deconvolution of two-source mixed gene expressions

Niya Wang[1], Eric P. Hoffman[2], Robert Clarke[3], Zhen Zhang[4], David M. Herrington[5], Ie-Ming Shih[4], Douglas A. Levine[6], Guoqiang Yu[1], Jianhua Xuan[1] and Yue Wang[1,†]

[1]Department of Electrical and Computer Engineering, Virginia Tech, Arlington, VA 22203, USA; [2]Research Center for Genetic Medicine, Children's National Medical Center, Washington, DC 20010, USA; [3]Lombardi Comprehensive Cancer Center, Georgetown University, Washington, DC 20057, USA; [4]Department of Pathology, Johns Hopkins University, Baltimore, MD 21231, USA; [5]Department of Internal Medicine, Wake Forest University, Winston-Salem, NC 27157, USA; [6]Department of Surgery, Memorial Sloan-Kettering Cancer Center, New York, NY 10021, USA

**Contact**: yuewang@vt.edu[†]

**Supplementary information**: Supplementary data are available at *Bioinformatics* online.

## 1 INTRODUCTION

Tissue heterogeneity is a major confounding factor in studying individual populations that cannot be resolved directly by global profiling (Hoffman, et al., 2004). Experimental solutions to mitigate tissue heterogeneity are expensive, time consuming, inapplicable to existing data, and may alter the original gene expression patterns (Kuhn, et al., 2011; Shen-Orr, et al., 2010). Alternatively, various *in silico* methods perform basically a supervised deconvolution based on either externally-obtained constituent proportions (Shen-Orr, et al., 2010; Stuart, et al., 2004) or previously-acquired cell-specific signatures (Kuhn, et al., 2011; Lu, et al., 2003).

In the earlier issues of this journal, a few articles have reported semi-supervised methods that were specifically focused on dissecting two-source mixed gene expressions. Gosink *et al.* used (known) expression data from a single cell type to determine the proportion (and subsequently expression profile) of each cell type in a heterogeneous sample (Gosink, et al., 2007). This method detects the minimum of a proportion that provides a good estimate in noiseless or simulation data. Built upon this work, Clarke *et al.* developed a geometry-based method that provides a more accurate estimate of this minimum in noisy real data and can be applied in situations where one or multiple heterologous samples are available (Clarke, et al., 2010). These methods assume a linear mixture of log-transformed expression levels that has recently been shown to be invalid (Zhong and Liu, 2012). Ahn *et al.* proposed a statistical approach for deconvolving linearly mixed cancer transcriptomes in individual samples under various raw measured data scenarios (Ahn, et al., 2013). This practical (most likely) solution again requires prior knowledge of gene expression of one tissue type from multiple similar samples and applies various heuristics.

Here we ask whether it is possible to deconvolute two-source mixed expressions (estimating both proportions and cell-specific profiles) from two or more heterogeneous samples without requiring any aforementioned prior knowledge (Wang, 2004). Supported by a well-grounded mathematical framework, we argue that both constituent proportions and cell-specific expressions can be estimated in a completely unsupervised mode when cell-specific marker genes exist, which do not have to be known a priori, for each of constituent cell types. Fundamental to the success of our approach is a geometric exploitation of cell/condition-specific marker genes and expression non-negativity. Specifically, we show that (1) the scatter plot of mixed expressions is a compressed version of cell-specific expressions; (2) the resident genes on the two radii of scatter sector are the cell-specific marker genes; (3) the radius vectors defined by the cell-specific marker genes are the column vectors of the mixing matrix; and (4) the rank of between-tissue differentially expressed genes is mixing-invariant. We demonstrate the performance of unsupervised deconvolution on both simulation and real gene expression data, together with perspective discussions.

## 2 THEORY AND METHOD

### 2.1 Roadmap of unsupervised deconvolution

We adopt the linear latent variable model of raw measured expression data (Zhong and Liu, 2012), given by (bold font indicates column vectors)

$$\begin{bmatrix} x_{\text{sample1}}(i) \\ x_{\text{sample2}}(i) \end{bmatrix} = \begin{bmatrix} a_{11} & a_{12} \\ a_{21} & a_{22} \end{bmatrix} \begin{bmatrix} s_{\text{tissue1}}(i) \\ s_{\text{tissue2}}(i) \end{bmatrix}, \text{ or } \mathbf{x}(i) = \mathbf{a}_1 s_{\text{tissue1}}(i) + \mathbf{a}_2 s_{\text{tissue2}}(i), \quad (1)$$

where $s_{\text{tissue1}}(i)$ and $s_{\text{tissue2}}(i)$ are the gene expression values for pure tissues 1-2, and $x_{\text{sample1}}(i)$ and $x_{\text{sample2}}(i)$ are the gene expression values for heterogeneous samples 1-2, for genes $i = 1,\ldots,n$, respectively; and $a_{jk}$ are the mixing proportions with $a_{11} + a_{12} = a_{21} + a_{22}$ (after signal normalization). We further adopt the concept of cell-specific marker genes (MG) (Gosink, et al., 2007; Kuhn, et al., 2011; Wang, 2004), i.e., genes whose expression is highly and exclusively enriched in a particular cell population in the context of interest, or mathematically $\mathbf{s}(i_{\text{MG1}}) \approx [\alpha_i \ 0]^T$ and $\mathbf{s}(i_{\text{MG2}}) \approx [0 \ \beta_i]^T$. Since raw measured gene expression values s are non-negative, when cell-specific marker genes exist for each cell type, the linear latent variable model (1) is identifiable using two or more mixed expressions, as we will elaborate via the following theorems and their formal proofs (see **Fig. 1** for geometric illustration).

**Theorem 1 (Scatter compression)**. *Suppose that pure tissue expressions are non-negative and* $\mathbf{x}(i) = \mathbf{a}_1 s_{\text{tissue1}}(i) + \mathbf{a}_2 s_{\text{tissue2}}(i)$ *where* $\mathbf{a}_1$ *and* $\mathbf{a}_2$ *are linearly* *independent, then, the scatter plot of mixed expressions is compressed into a scatter sector whose two radii coincide with* $\mathbf{a}_1$ *and* $\mathbf{a}_2$.

*Proof of theorem 1.* Since $\mathbf{a}_1$ and $\mathbf{a}_2$ are linearly independent, without loss of generality, we assume that

$$\frac{a_{12}}{a_{22}} < \frac{a_{11}}{a_{21}}, \text{ i.e., } a_{12}a_{21} < a_{11}a_{22}.$$

Multiply both sides by $s_{\text{tissue2}}(i)$ and add $a_{11}a_{21}s_{\text{tissue1}}(i)$ to both sides, since raw measured expressions are non-negative, we have

$$a_{11}a_{21}s_{\text{tissue1}}(i) + a_{12}a_{21}s_{\text{tissue2}}(i) < a_{11}a_{21}s_{\text{tissue1}}(i) + a_{11}a_{22}s_{\text{tissue2}}(i).$$

Simple mathematical reorganizations lead to

$$\left(a_{11}s_{\text{tissue1}}(i) + a_{12}s_{\text{tissue2}}(i)\right)a_{21} < a_{11}\left(a_{21}s_{\text{tissue1}}(i) + a_{22}s_{\text{tissue2}}(i)\right),$$

$$\rightarrow \frac{a_{11}s_{\text{tissue1}}(i) + a_{12}s_{\text{tissue2}}(i)}{a_{21}s_{\text{tissue1}}(i) + a_{22}s_{\text{tissue2}}(i)} < \frac{a_{11}}{a_{21}}.$$

Since $\mathbf{x}(i) = \mathbf{a}_1 s_{\text{tissue1}}(i) + \mathbf{a}_2 s_{\text{tissue2}}(i)$, we have

$$\frac{x_{\text{sample1}}(i)}{x_{\text{sample2}}(i)} < \frac{a_{11}}{a_{21}}.$$

Using a similar strategy, we can show

$$\frac{a_{12}}{a_{22}} < \frac{x_{\text{sample1}}(i)}{x_{\text{sample2}}(i)},$$

that readily completes the proof. Q.E.D.

**Theorem 2 (Unsupervised identifiability).** *Suppose that pure tissue expressions are non-negative and cell-specific marker genes exist for each constituting tissue type, and* $\mathbf{x}(i) = \mathbf{a}_1 s_{\text{tissue1}}(i) + \mathbf{a}_2 s_{\text{tissue2}}(i)$ *where* $\mathbf{a}_1$ *and* $\mathbf{a}_2$ *are linearly independent, then, the two radii of the scatter sector of mixed expressions coincide with* $\mathbf{a}_1$ *and* $\mathbf{a}_2$ *that can be readily estimated from marker gene expression values with appropriate rescaling.*

*Proof of theorem 2.* Based on the definition of cell-specific marker genes, i.e., $\mathbf{s}(i_{\text{MG1}}) = [\alpha_i\ 0]^T$ and $\mathbf{s}(i_{\text{MG2}}) = [0\ \beta_i]^T$, and the existence of cell-specific marker genes for all constituting tissue types, we have

$$\mathbf{x}(i_{\text{MG1}}) = \mathbf{a}_1 \alpha_i, \ \mathbf{x}(i_{\text{MG2}}) = \mathbf{a}_2 \beta_i.$$

By the conclusion of Theorem 1, we complete the proof. Q.E.D.

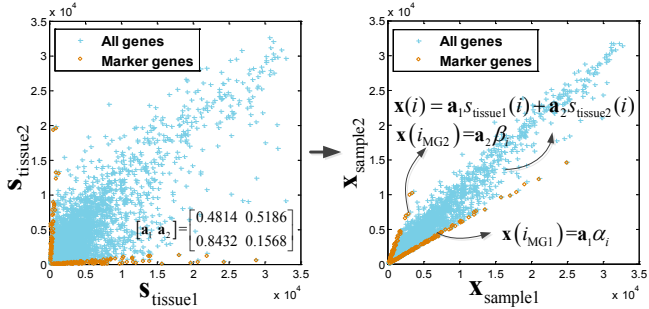

**Figure 1.** Geometric and mathematical description of the mixing process.

**Corollary 1 (Invariance of differential expression).** *Suppose that pure tissue expressions are non-negative and* $\mathbf{x}(i) = \mathbf{a}_1 s_{\text{tissue1}}(i) + \mathbf{a}_2 s_{\text{tissue2}}(i)$ *where* $\mathbf{a}_1$ *and* $\mathbf{a}_2$ *are linearly independent, then, the rank of between-tissue differentially expressed genes is mixing-invariant.*

*Proof of corollary 1.* Without loss of generality, we assume that

$$\frac{a_{12}}{a_{22}} < \frac{a_{11}}{a_{21}}, \text{ i.e., } a_{12}a_{21} < a_{11}a_{22},$$

and

$$\frac{s_{\text{tissue1}}(j)}{s_{\text{tissue2}}(j)} < \frac{s_{\text{tissue1}}(i)}{s_{\text{tissue2}}(i)}, \text{ i.e., } s_{\text{tissue1}}(j)s_{\text{tissue2}}(i) < s_{\text{tissue1}}(i)s_{\text{tissue2}}(j).$$

Since $\mathbf{a}_1$ and $\mathbf{a}_2$ are linearly independent and $a_{12}a_{21} < a_{11}a_{22}$, multiply both sides by $(a_{11}a_{22} - a_{12}a_{21})$, and add $a_{11}a_{21}s_{\text{tissue1}}(i)s_{\text{tissue1}}(j)$ and $a_{12}a_{22}s_{\text{tissue2}}(i)s_{\text{tissue2}}(j)$ to both sides, we have

$$a_{11}a_{21}s_{\text{tissue1}}(i)s_{\text{tissue1}}(j) + (a_{11}a_{22} - a_{12}a_{21})s_{\text{tissue1}}(j)s_{\text{tissue2}}(i)$$
$$+ a_{12}a_{22}s_{\text{tissue2}}(i)s_{\text{tissue2}}(j)$$
$$<$$
$$a_{11}a_{21}s_{\text{tissue1}}(i)s_{\text{tissue1}}(j) + (a_{11}a_{22} - a_{12}a_{21})s_{\text{tissue1}}(i)s_{\text{tissue2}}(j)$$
$$+ a_{12}a_{22}s_{\text{tissue2}}(i)s_{\text{tissue2}}(j).$$

Simple mathematical reorganizations lead to

$$a_{11}a_{21}s_{\text{tissue1}}(i)s_{\text{tissue1}}(j) + a_{12}a_{21}s_{\text{tissue1}}(i)s_{\text{tissue2}}(j)$$
$$+ a_{11}a_{22}s_{\text{tissue1}}(j)s_{\text{tissue2}}(i) + a_{12}a_{22}s_{\text{tissue2}}(i)s_{\text{tissue2}}(j)$$
$$<$$
$$a_{11}a_{21}s_{\text{tissue1}}(i)s_{\text{tissue1}}(j) + a_{12}a_{21}s_{\text{tissue1}}(j)s_{\text{tissue2}}(i)$$
$$+ a_{11}a_{22}s_{\text{tissue1}}(i)s_{\text{tissue2}}(j) + a_{12}a_{22}s_{\text{tissue2}}(i)s_{\text{tissue2}}(j),$$

$$\rightarrow \frac{a_{11}s_{\text{tissue1}}(j) + a_{12}s_{\text{tissue2}}(j)}{a_{21}s_{\text{tissue1}}(j) + a_{22}s_{\text{tissue2}}(j)} < \frac{a_{11}s_{\text{tissue1}}(i) + a_{12}s_{\text{tissue2}}(i)}{a_{21}s_{\text{tissue1}}(i) + a_{22}s_{\text{tissue2}}(i)}.$$

Since $\mathbf{x}(i) = \mathbf{a}_1 s_{\text{tissue1}}(i) + \mathbf{a}_2 s_{\text{tissue2}}(i)$, we complete the proof with

$$\frac{s_{\text{tissue1}}(j)}{s_{\text{tissue2}}(j)} < \frac{s_{\text{tissue1}}(i)}{s_{\text{tissue2}}(i)} \rightarrow \frac{x_{\text{sample1}}(j)}{x_{\text{sample2}}(j)} < \frac{x_{\text{sample1}}(i)}{x_{\text{sample2}}(i)}. \quad \text{Q.E.D.}$$

From Theorem 2, there exists a mathematical solution uniquely identifying the linear latent variable model (1) based on two/more mixed expressions: under a noise-free scenario, we can (in principle) directly estimate $\mathbf{a}_1$ and $\mathbf{a}_2$ by locating the two radii that most tightly enclose the scatter sector of mixed expressions. Moreover, Corollary 1 allows for between-tissue differential analysis from mixed expressions without requiring deconvolution.

## 2.2 Algorithm and evaluation criteria

So far, we have described the theoretical roadmap for unsupervised deconvolution of two-source mixed expressions. We now complete the description of our algorithm by considering the identification of marker genes or mixing matrix, and its application to data deconvolution.

Although the mixing matrix can be estimated using only one marker gene per tissue type, a more accurate solution, with practical applicability, is to estimate $\mathbf{a}_1$ and $\mathbf{a}_2$ using multiple marker genes. Our unsupervised deconvolution begins by detecting the cell-specific marker genes directly from mixed expressions, in which the differential analysis of gene expressions is performed on all genes. Mathematically, MG is defined as an index set

$$\text{MG} = \left\{ i \middle| \left[ k_{\max} - \varepsilon \leq \frac{x_{\text{sample2}}(i)}{x_{\text{sample1}}(i)} \leq k_{\max} \right] \cup \left[ k_{\min}^{-1} - \varepsilon \leq \frac{x_{\text{sample1}}(i)}{x_{\text{sample2}}(i)} \leq k_{\min}^{-1} \right] \right\}, \quad (2)$$

where $k_{\max}$ and $k_{\min}$ are the maximum and minimum ratios of $x_{\text{sample2}}(i)$ over $x_{\text{sample1}}(i)$ across all $i$, respectively; and $\varepsilon$ is a pre-fixed positive small real number. To obtain a reliable set of marker gene indices, some pre-processing steps are required, including mode/mean-based normalization and removal of minimally-expressed and outlier genes.

On the basis of the expression values of detected cell-specific marker genes, the mixing matrix is estimated using standardized sample averages,

$$\hat{\mathbf{a}}_1 = \frac{1}{n_{\text{MG1}}} \sum_{i \in \text{MG1}} \frac{\mathbf{x}(i)}{\|\mathbf{x}(i)\|}, \ \hat{\mathbf{a}}_2 = \frac{1}{n_{\text{MG2}}} \sum_{i \in \text{MG2}} \frac{\mathbf{x}(i)}{\|\mathbf{x}(i)\|}, \quad (3)$$

where MG1 and MG2 are the index sets of marker genes for tissue types 1 and 2, respectively; $n_{\text{MG1}}$ and $n_{\text{MG2}}$ are the numbers of marker genes for tissue types 1 and 2, respectively; and $\|.\|$ denotes the vector norm (L1 or L2). The resulting $\hat{\mathbf{a}}_1$ and $\hat{\mathbf{a}}_2$ are then used to deconvolute the mixed expressions into cell-specific profiles via matrix inversion techniques.

**Unsupervised deconvolution algorithm:**
1) Normalize gene expression profile using global mean/mode;
2) Remove minimally-expressed genes whose norm is less than a pre-fixed positive small real number $\delta$, and outlier genes whose norm is bigger than a pre-fixed positive large real number $\gamma$;

3) Detect the indices of cell-specific marker genes, for each of the constituting tissue types, according to (2);
4) Estimate mixing matrix according to (3), normalized to proportions;
5) Estimate cell-specific expression profiles using mixed expressions and matrix inversion technique(s).

We use four complementary evaluation criteria and known ground truth to assess the performance of the proposed unsupervised deconvolution method. To assess the accuracy of tissue proportion estimates, in addition to classic correlation coefficient, we adopt the E1 criterion given by

$$E1 = \sum_{i=1}^{2}\left(\sum_{j=1}^{2}\frac{|p_{ij}|}{\max_k |p_{ik}|}-1\right) + \sum_{j=1}^{2}\left(\sum_{i=1}^{2}\frac{|p_{ij}|}{\max_k |p_{kj}|}-1\right), \quad (4)$$

where $p_{ij}$ is the *ij*th element of the matrix $[\hat{\mathbf{a}}_1\ \hat{\mathbf{a}}_2]^{-1}[\mathbf{a}_1\ \mathbf{a}_2]$ with $\hat{\mathbf{a}}_1$ and $\hat{\mathbf{a}}_2$ being the estimated column vectors of mixing matrix. Note that E1 is invariant to permutation or scaling and E1=0 when the estimation is perfect. To assess the accuracy of estimated cell-specific expression patterns, we calculate the correlation coefficient between the estimated expression profile and ground truth over 'marker genes' and 'all genes' respectively. Moreover, to assess the membership (and rank) match (and mismatch) between the marker genes detected from pure versus mixed expressions, we utilize Venn diagrams, together with Spearman's rank correlation coefficient.

More details on algorithm, parameter settings, and alternative schemes, are included in the supplementary information.

## 3 EXPERIMENTAL RESULTS

### 3.1 Validation on cell line expression data

We first considered numerical mixtures of two human cell line expressions, a situation in which all factors are known and linear mixture model is ideal. We reconstituted mixed expressions by multiplying the measured cell line expressions by the proportion of the tissue subset in a given heterogeneous sample (**Fig. 1**). We detected the cell-specific marker genes solely based on the reconstituted expression mixtures and accordingly obtained a highly accurate estimate of the mixing matrix with E1=0.0295 (and a correlation coefficient of 0.99) (**Table 1**). For each cell line, a comparison of the estimated expression profile of each type to the measured expression pattern in the pure cell line showed an almost perfect correlation with an average correlation coefficient of 0.99, indicating that we could accurately deconvolute the mixed expressions into constituent expression patterns in a completely unsupervised way.

**Table 1.** Identification of linear mixture model using numerical mixtures of two breast cancer cell line expressions.

| Sample/Tissue | MCF7 (breast cancer) (assigned/estimated) | HS27 (fibroblasts) (assigned/estimated) |
|---|---|---|
| Sample 1 | 0.4814/0.4771 | 0.5186/0.5229 |
| Sample 2 | 0.8432/0.8336 | 0.1568/0.1664 |

Next, we tested our method on biologically mixed expressions from two breast cancer cell lines. The mRNA extracted from the individual cell lines are mixed with pre-specified proportions before subsequent procedures including amplification and microarray experiment (**Table 2**). Such mixtures mimic the actual biological samples with varying relative abundances of the constituent subsets from one another (**Fig. 2**) (Kuhn, et al., 2011; Shen-Orr, et al., 2010).

**Table 2.** Identification of linear mixture model using biological mixtures of two breast cancer cell line expressions.

| Sample/Tissue | MCF7 (breast cancer) (assigned/estimated) | HS27 (fibroblasts) (assigned/estimated) |
|---|---|---|
| Sample 1 | 0.75/0.7622 | 0.25/0.2378 |
| Sample 2 | 0.25/0.2424 | 0.75/0.7576 |

The proposed method again accurately estimated the mixing proportions with E1=0.0778 (and a correlation coefficient of 0.99) (**Table 2** and **Fig. 2**), and cell-specific expression patterns with an average correlation coefficient of 0.99 between the estimated expression profile of each type to the measured expression pattern in the pure cell line (**Fig. 3**). The high correlation that we achieved between the estimated proportions/tissue-expressions and ground truth suggests that unsupervised deconvolution of tissue-specific expression profiles from two-source heterogeneous samples using a linear model should yield accurate expression estimates for most genes.

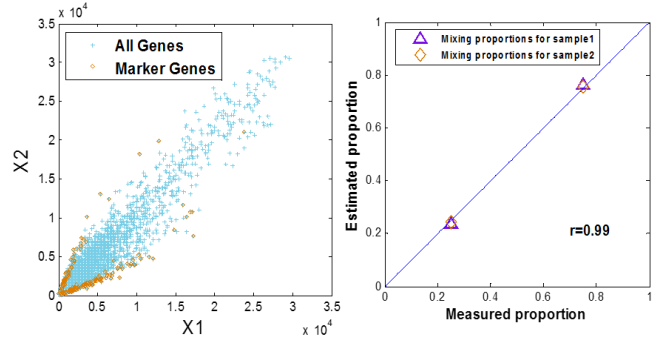

**Figure 2.** Scatter plot of biological mixing and blind model identification.

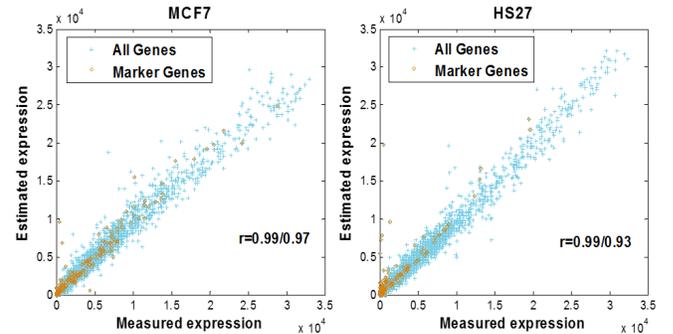

**Figure 3.** Highly correlated scatter plots between the estimated and measured pure cell expression profiles (over 'marker' and 'all' genes).

The theoretical roadmap also enables the extended detection of differentially-expressed genes beyond marker genes aimed at maximizing the information obtainable from mixed expressions. To assess the specificity and sensitivity of detecting differentially expressed genes without deconvolution, we compared the ranked index subsets of differentially expressed genes between samples to a 'gold standard' set of differentially expressed genes identified from the pure cell line measurements, on both numerical and biological mixtures of two breast cancer cell line expressions. In addition to the Venn diagram and Spearman's rank correlation coefficient ($r_{\text{rank}}$=0.92), receiver operating characteristics curve analysis showed that the detection of differentially expressed genes based on mixed expressions (Corollary 1) to be both highly specific and sensitive with an area under the curve of 0.85 (supplementary information).

### 3.2 Analysis of benchmark expression data

As an example for the purpose of comparison, we also analyzed the same public benchmark gene expression dataset (AFFY) used by Ahn *et al.* (Ahn, et al., 2013). This dataset consists of biologically mixed heterogeneous samples with varying proportions of human brain and heart tissues. We selected 12 samples with brain/heart proportion ratios of 0/100% (3 samples), 25%/75% (3 samples), 75%/25% (3 samples) and 100%/0 (3 samples). In contrast to the semi-supervised methods that all require prior knowledge of gene expression of one tissue type (Ahn, et al., 2013; Clarke, et al., 2010; Gosink, et al., 2007), the 6 pure tissue samples were not used in any step of our proposed algorithm but simply served as the truth for assessment.

For each proportion ratio, consistent with routine practice, we take the average of the 3 replicates (with the same proportion ratio) as the mixed/observed expression profile to be analyzed by the algorithm. As aforementioned, in addition to accurate signal normalization (Wang, et al., 2002), to maintain data quality and computational efficiency, we selected a subset of genes in the subsequent analyses by excluding minimally-expressed and outlier genes (supplementary information) (Ahn, et al., 2013). This provided us with about 11000 probe sets across all samples.

**Table 3.** Unsupervised estimation of unknown tissue proportions on AFFY brain-heart mixture dataset.

| Sample/Tissue | Brain (assigned/estimated) | Heart (assigned/estimated) |
|---|---|---|
| Sample 1 | 0.25/0.2342 | 0.75/0.7658 |
| Sample 2 | 0.75/0.7622 | 0.25/0.2378 |

With pre-processed raw measured data, we first evaluated how well the proposed method estimated tissue proportions in this dataset (**Table 3**). Without using any knowledge of either tissue-specific expression or constituent proportions, as in other methods (Ahn, et al., 2013; Clarke, et al., 2010; Gosink, et al., 2007; Kuhn, et al., 2011), our algorithm accurately estimated the unknown tissue proportions with a correlation coefficient of 0.99 (E1=0.1247), as compared with a correlation coefficient of 0.98 produced by the semi-supervised method on the same dataset (**Fig. 4**) (Ahn, et al., 2013).

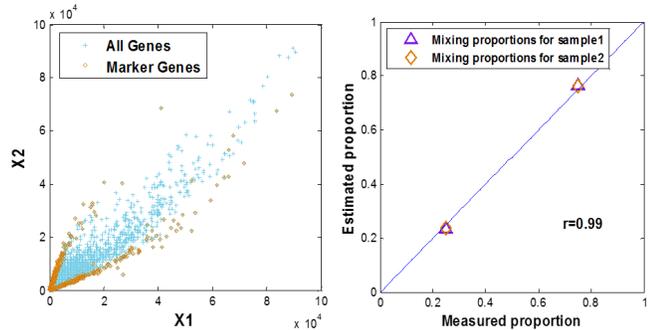

**Figure 4.** Scatter plot of brain-heart mixtures and proportion estimates.

Next, we examined how well the proposed method estimated tissue-specific expression patterns in this dataset. As shown in **Fig. 5**, the proposed method accurately and blindly estimated the gene expressions of pure brain and pure heart tissues, with a correlation coefficient of 0.96-0.99 between the estimated mean tissue expression levels and measured mean pure tissue expression levels, as compare to a correlation coefficient of 0.88-0.95 produced by the semi-supervised method on the same dataset. These results suggest that this unsupervised deconvolution method is able to accurately deconvolute two-source mixed expressions (estimating both proportions and cell-specific profiles) from two or more heterogeneous samples.

Detailed information on additional experimental results (tables, figures, datasets) are included in the supplementary information.

## 4 DISCUSSION

In this letter, we presented a feasible roadmap for unsupervised deconvolution of two-source mixed expressions, supported by the newly proved theorems under realistic conditions and experimental tests on real gene expression data. One important advantage of this unsupervised deconvolution approach lies in its unique and proven ability to detect cell-specific marker genes and estimate constituent proportions directly from mixed expressions when the relevant prior knowledge is either unreliable or unavailable. This is significant, in relation to semi-supervised methods, since it is well-known that (1) cell-specific marker genes (membership and expression) are condition-specific and (2) the total amount of mRNA from the same volume of cancer cells is much higher than that of normal cells (due to unknown tumor ploidy) (Clarke, et al., 2010).

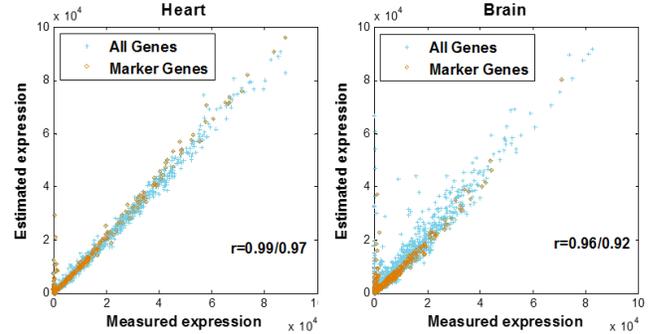

**Figure 5.** Estimation of tissue-specific gene expressions from AFFY: scatter plots comparing deconvolved mean brain/heart tissue expression values with measured mean pure brain/heart tissue expression values.

We foresee a variety of extensions to the concepts and strategies in the proposed method. For example, with further development, intratumor heterogeneity can be revealed in terms of hidden subclonal marker genes and subclonal repopulation dynamics.

There is also a possible way to estimate the marker expression profiles for individual samples (Ahn, et al., 2013). Rewrite (1) as

$$\begin{bmatrix} x_{\text{sample1}}(i) \\ x_{\text{sample2}}(i) \end{bmatrix} = \begin{bmatrix} a_{11} & a_{12} \\ a_{21} & a_{22} \end{bmatrix} \begin{bmatrix} s_{\text{tissue1}}(i) + \Delta s_{\text{tissue1,sample1}}(i) \\ s_{\text{tissue2}}(i) + \Delta s_{\text{tissue2,sample2}}(i) \end{bmatrix}, \quad (5)$$

where $\Delta s_{\text{tissue1,sample1}}(i)$ and $\Delta s_{\text{tissue2,sample2}}(i)$ are the sample-specific variations in pure tissues. Then, for marker genes, we have

$$\begin{bmatrix} x_{\text{sample1}}(i_{\text{MG}j}) \\ x_{\text{sample2}}(i_{\text{MG}j}) \end{bmatrix} = \begin{bmatrix} a_{1j}\left(s_{\text{tissue}j}(i_{\text{MG}j}) + \Delta s_{\text{tissue}j,\text{sample1}}(i_{\text{MG}j})\right) \\ a_{2j}\left(s_{\text{tissue}j}(i_{\text{MG}j}) + \Delta s_{\text{tissue}j,\text{sample2}}(i_{\text{MG}j})\right) \end{bmatrix}, \quad (6)$$

where $j$ is the tissue type index. According to (3), $\hat{a}_{1j}$ and $\hat{a}_{2j}$ are obtained via some form of 'averaging' over tissue-specific marker genes, where for each sample we may reasonably assume

$$\frac{1}{n_{\text{MG}j}} \sum_{i \in \text{MG}j} \frac{\Delta s_{\text{tissue}j,\text{sample}}(i)}{\|\mathbf{x}(i)\|} \approx 0. \quad (7)$$

Denote $s_{\text{tissue}j,\text{sample}k}(i_{\text{MG}j}) = s_{\text{tissue}j}(i_{\text{MG}j}) + \Delta s_{\text{tissue}j,\text{sample}k}(i)$, we have

$$s_{\text{tissue}j,\text{sample}k}(i_{\text{MG}j}) = x_{\text{sample}k}(i_{\text{MG}j}) / a_{kj}, \quad (8)$$

for each of $k$ and $j$, where $k$ is the sample index.

*Funding*: National Institutes of Health, under Grants NS29525, CA149147, CA160036, HL111362, in part.

*Conflict of Interest*: none declared.